\documentclass{article}
\usepackage[utf8]{inputenc}
\usepackage[T1]{fontenc}
\usepackage{hyperref}
\usepackage{url}
\usepackage{booktabs}
\usepackage{amsfonts}
\usepackage{amsmath}
\usepackage{nicefrac}
\usepackage{microtype}
\usepackage{graphicx}
\usepackage{xcolor}
\usepackage{amsfonts}
\usepackage{lipsum}
\usepackage{booktabs}
\usepackage{multirow}
\usepackage{graphicx}
\usepackage{subcaption}
\usepackage{array}
\usepackage{bbold}
\usepackage{float}
\usepackage[margin=1.5in]{geometry}
\usepackage{amssymb}

\title{
   Enhancing Robotic Manipulation: Harnessing the Power of Multi-Task Reinforcement Learning\\ and Single Life Reinforcement Learning in Meta-World\\
  \vspace{1em}
}

\author{}
\date{}

\begin{document}

\maketitle

\vspace{-2cm}
\begin{center}
    Ghadi Nehme\footnote{Stanford University}, Ishan Sabane\footnotemark[\value{footnote}], Tejas Y. Deo\footnotemark[\value{footnote}]
\end{center}
\vspace{0.5cm}
\begin{abstract}
    
At present, robots typically require extensive training to successfully accomplish a single task. However, to truly enhance their usefulness in real-world scenarios, robots should possess the capability to perform multiple tasks effectively. To address this need, various multi-task reinforcement learning (RL) algorithms have been developed \cite{yu2020meta}, including multi-task proximal policy optimization (PPO), multi-task trust region policy optimization (TRPO), and multi-task soft-actor critic (SAC). Nevertheless, these algorithms demonstrate optimal performance only when operating within an environment or observation space that exhibits a similar distribution. In reality, such conditions are often not the norm, as robots may encounter scenarios or observations that differ from those on which they were trained.

For instance, if a robot is trained to pick and place a sphere from a specific set of locations, it should also possess the capability, during testing, to successfully pick and place a cube from different positions, at least those that are in close proximity to the trained positions. Addressing this challenge, algorithms like Q-Weighted Adversarial Learning (QWALE) \cite{NEURIPS2022_5ec4e93f} attempt to tackle the issue by training the base algorithm (generating prior data) solely for a particular task, rendering it unsuitable for generalization across tasks.

And so, the aim of this research project is to enable a robotic arm to successfully execute seven distinct tasks within the Meta World environment. These tasks encompass Pick and Place, Window Open and Close, Drawer Open and Close, Push, and Button Press. To achieve this, a multi-task soft actor critic (MT-SAC) is employed to train the robotic arm. Subsequently, the trained model will serve as a source of prior data for the single-life RL algorithm. The effectiveness of this MT-QWALE algorithm will be assessed by conducting tests on various target positions (novel positions), i.e., for example, different window positions in the Window Open and Close task.

At the end, a comparison is provided between the trained MT-SAC and the MT-QWALE algorithm where the MT-QWALE performs better. An ablation study demonstrates that MT-QWALE successfully completes tasks with slightly more number of steps even after hiding the final goal position.


    
\end{abstract}

\newpage


\section{Introduction}

Reinforcement learning (RL) has made significant strides in various domains, including robotic manipulation, Atari games, etc. However, existing state-of-the-art RL methods require significantly more experience than humans to acquire even a single narrowly-defined skill. To make robots genuinely useful in realistic environments, it is crucial to develop algorithms that can reliably and efficiently learn a wide range of skills. Fortunately, in specific domains like robotic manipulation or locomotion, many tasks share common underlying structures, which can be leveraged to learn related tasks more efficiently.

For example, most robotic manipulation tasks involve actions such as grasping, or moving objects within the workspace. Although current methods can learn individual skills like pick and place or hanging a mug or hammering a screw, we require algorithms that can efficiently learn and utilize shared structures across multiple related tasks. These algorithms should be able to learn new skills quickly by leveraging the learned structure, such as screwing a jar lid or hanging a bag. Recent advancements in machine learning have demonstrated exceptional generalization capabilities in domains like speech, implying that similar generalizations should be achievable in RL settings encompassing diverse tasks.

Emerging approaches in meta-learning and multi-task reinforcement learning show promise in addressing this challenge. Multi-task RL methods aim to learn a single policy capable of efficiently solving multiple tasks, surpassing the performance of learning individual tasks independently. On the other hand, meta-learning methods train on a multitude of tasks and optimize for rapid adaptation to new tasks.

In the context of multi-task and meta-RL settings, it is crucial to address the challenge of agent failure when confronted with new and unseen situations. While a multi-task RL algorithm may exhibit excellent performance across a set of 50 tasks, it can still struggle significantly when faced with an observation space different from its training data. This limitation becomes evident in scenarios where a robotic manipulator, trained to open drawers at specific locations, encounters a completely different drawer position during testing. Similarly, for search-and-rescue disaster relief robots navigating through buildings, the possibility of encountering unfamiliar obstacles further highlights the need for adaptability and effective handling of new circumstances.

Traditional RL algorithms aim to optimize the solution for a fixed set of tasks, without considering on-the-fly adaptations without human intervention. On the other hand, the Single Life RL (SLRL) setting encourages the autonomous and successful resolution of novel scenarios within a single trial. However, the current SLRL agents are primarily designed to handle novelties within a single environment, such as pick and place tasks or opening drawers.

\section{Related Work} 

Many prior works and frameworks have been developed to test the effectiveness of the multi-task RL framework. In Meta-world: A benchmark and evaluation for multi-task and meta reinforcement learning \cite{yu2020meta}, a framework has been developed to evaluate the capabilities of current multi-task and meta-reinforcement learning methods and make it feasible to design new algorithms that actually generalize and adapt quickly on meaningfully distinct tasks. Evaluation protocols and task suites in Metaworld are broad enough to enable this sort of generalization while containing sufficient shared structure for generalization to be possible.

You Only Live Once: Single-Life Reinforcement Learning \cite{NEURIPS2022_5ec4e93f} introduces a challenge where an agent must successfully complete a task within a single trial without human intervention, adapting to novel situations. Standard episodic RL algorithms struggle in this setting, prompting the proposal of Q-weighted adversarial learning (QWALE). QWALE leverages distribution matching and prior experience to guide the agent in novel states, resulting in 20-60\% higher success rates in single-life continuous control problems. This work provides valuable insights into autonomously adapting to unfamiliar situations, highlighting the effectiveness of distribution matching-based methods in SLRL. The major drawback of this approach is that the prior data for each task is collected by training a task-specific algorithm like SAC and thus is not able to generalize well on multiple tasks.

Multi-task reinforcement learning with soft modularization \cite{yang2020multi} addresses the challenge of multi-task learning in reinforcement learning, emphasizing the complexity of optimizing shared parameters across tasks. To overcome this, an explicit modularization technique is introduced, utilizing a routing network to reconfigure the base policy network for each task. The task-specific policy employs soft modularization, combining possible routes for sequential tasks. Experimental results on robotics manipulation tasks demonstrate significant improvements in sample efficiency and performance compared to strong baselines. Even though the performance was significantly improved due to the soft modularization approach, the agent will still fail badly when there is some distribution shift in the observations. 

Actor-mimic: Deep multitask and transfer reinforcement learning \cite{parisotto2015actor} introduces a novel approach to enable agents to learn and transfer knowledge across multiple tasks. By utilizing deep reinforcement learning and model compression techniques, the proposed "Actor-Mimic" method trains a single policy network guided by expert teachers. The learned representations demonstrate the ability to generalize to new tasks without prior expert guidance, speeding up learning in novel environments. Atari games serve as a testing ground to showcase the effectiveness of the approach. Even though the approach is computationally efficient, it does not focus on the behavior of the agent on the same task with novel scenarios/observations. 

Comparing task simplifications to learn closed-loop object picking using deep reinforcement learning \cite{breyer2019comparing} compares reinforcement learning-based approaches for object picking in unstructured environments with a robotic manipulator. It shows that policies learned through curriculum learning and sparse rewards achieve similar success rates to those initialized on simplified tasks, with successful transfer to the real robot. The robotics arm is controlled by moving the end-effector by a certain displacement vector and changing the torque that the grippy fingers should apply. Thus, the robotics arm is controlled with inverse kinematics.

\section{Background}

For each task $\mathcal{T}$, we examine a finite horizon Markov decision process (MDP) where $M$ tasks exist in total. These MDPs can be denoted as $(S, A, P, R, H, \gamma)$, where both the state $s \in S$ and action $a \in A$ are continuous. The stochastic transition dynamics are represented by $P(s_{t+1}\mid s_t, a_t)$, and the reward function is denoted as $R(s_t, a_t)$. The horizon is defined as $H$, and $\gamma$ represents the discount factor. To represent the policy parameterized by $\phi$, we utilize $\pi_\phi(a_t\mid s_t)$, with the objective being to learn a policy that maximizes the expected return. In the context of multi-task RL, tasks are randomly selected from a distribution $p(\mathcal{T})$, and each task corresponds to a distinct MDP.

\subsection{Meta-World}

Meta-World presents an open-source simulated benchmark designed for meta-reinforcement learning and multi-task learning. It encompasses 50 unique robotic manipulation tasks that involve the interaction of a robotic arm with diverse objects possessing varying shapes, joints, and connectivity. Each task necessitates the robot to perform a combination of reaching, pushing, and grasping actions, depending on the specific task at hand.

\begin{figure}[H]
\begin{center}
\includegraphics[width = 0.8\textwidth]{ 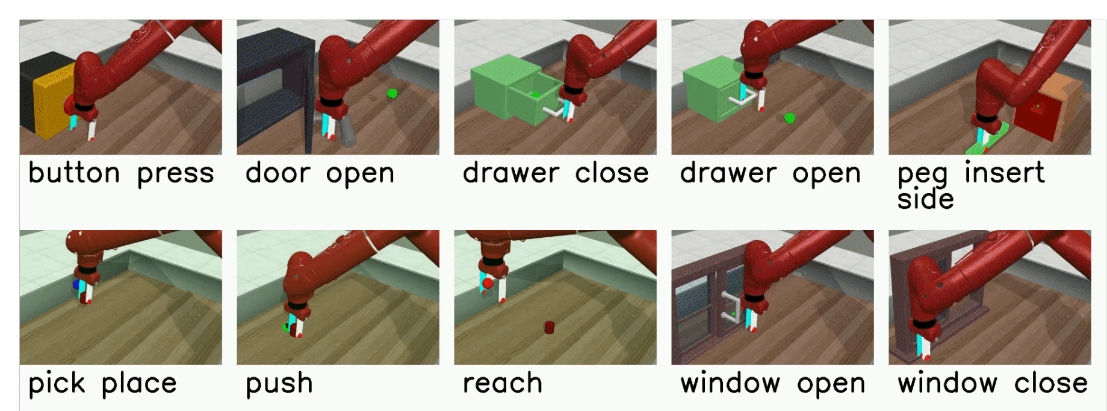}
\end{center}
   \caption{Multi-Task 10 (MT-10) Environment in Meta-World}
\label{fig:1}
\end{figure}

\subsubsection{Action Space}

The action space for Meta-World can be represented with the following vector:
\[\mathbf{a} = 
\begin{bmatrix}
\delta \mathbf{x} : \text{3D space of the end-effector} \\
\tau : \text{normalized torque that the gripper fingers should apply}
\end{bmatrix}
\in \mathbb{R}^{4}
\]
\\
In the reinforcement learning context, the action is the output of a trained agent.

\subsubsection{Observation Space}

The observation space for Meta-World can be represented with the following vector:
\[\mathbf{o} = 
\begin{bmatrix}
\mathbf{x}_E^{(i)} : \text{3D Cartesian coordinates of End-Effector}\\
\delta_G^{(i)}: \text{Measurement of how open the gripper is}\\
\mathbf{x}_1^{(i)} : \text{3D position of the first object}\\
\mathbf{\lambda}_1^{(i)} : \text{Quaternion of the first object}\\
\mathbf{x}_2^{(i)} : \text{3D position of the second object}\\
\mathbf{\lambda}_2^{(i)} : \text{Quaternion of the second object}\\
\mathbf{x}_{\text{Goal}} : \text{3D position of the goal}
\end{bmatrix}
\in \mathbb{R}^{39}, i\in \{t, t-1\}
\]

In the reinforcement learning context, the observation is the output of a trained agent.

\subsubsection{Reward Functions}

In Meta-World, reward functions have two key factors. First, tasks must be manageable by existing single-task reinforcement learning algorithms. This is crucial for evaluating multi-task and meta-reinforcement learning algorithms. Second, the reward functions should exhibit shared structure across tasks. This promotes the transfer of knowledge and skills between tasks, enhancing the effectiveness of multi-task and meta-reinforcement learning in Meta-World.

\subsection{Soft Actor-Critic for Reinforcement Learning}
This paper introduces Soft Actor-Critic (SAC) as the training method for policy optimization. SAC is a deep reinforcement learning approach that combines off-policy actor-critic techniques and aims to achieve task success while maximizing randomness in action selection. The parameterized soft Q-function, denoted as $Q_\theta(s_t, a_t)$, where $\theta$ represents the parameters, is considered. SAC involves optimizing three types of parameters: the policy parameters $\phi$, the parameters of the Q-function $\theta$, and a temperature parameter $\alpha$. The objective of policy optimization can be defined as follows:
\begin{equation}
    J_\pi(\phi) = \mathbb{E}_{s_t \sim \mathcal{D}}\left[\mathbb{E}_{a_t \sim \pi_\phi}\left[\alpha \log \pi_\phi (a_t \mid s_t) - Q_\theta (s_t,a_t)\right]\right]
\end{equation}

Here, $\alpha$ is a learnable temperature that serves as an entropy penalty coefficient. It can be learned to maintain the desired entropy level of the policy, using the following equation:

$$J(\alpha) = \mathbb{E}_{a_t \sim \pi_\phi}\left[-\alpha \log \pi_\phi (a_t \mid s_t) - \alpha \Tilde{\mathcal{H}}\right]$$

where $\Tilde{\mathcal{H}}$ represents the desired minimum expected entropy. If the optimization of $\log \pi_t(a_t\mid s_t)$ increases its value and reduces entropy, $\alpha$ is adjusted accordingly to increase the process.

\subsection{Multi-task Reinforcement Learning}
To extend SAC from single-task to multi-task scenarios, we introduce a single, task-conditioned policy $\pi(a\mid s, z)$, where $z$ represents a task embedding. The policy is optimized to maximize the average expected return across all tasks sampled from the distribution $p(\mathcal{T})$. The objective of policy optimization is defined as follows:

$$J_\pi(\phi) = \mathbb{E} _{\mathcal{T} \sim p(\mathcal{T})} [J_{\pi,\mathcal{T}} (\phi)],$$

where $J_{\pi,\mathcal{T}} (\phi)$ is directly adopted from Equation 1, incorporating the specific task $\mathcal{T}$. Similarly, for the Q-function, the objective is as follows:

$$J_Q(\theta) = \mathbb{E}_{\mathcal{T}\sim p(\mathcal{T})} \left[J_{Q,\mathcal{T}}(\theta)\right]$$

\subsection{Single Life Reinforcement Learning}

Single-life reinforcement learning enables autonomous task completion in a single trial, in the presence of a novel distribution shift, by leveraging prior data.

The prior data $\mathcal{D}_\text{prior}$ consists of transition data from transitions from source MDP $\mathcal{M}_\text{source}$. The agent interacts with the target MDP defined by $\mathcal{M}_\text{target}$ $(S, A, \Tilde{P}, R, \Tilde{\rho}, \gamma)$

The target MDP is assumed to have a novelty that is not present in the source MDP such as different dynamics or differences in the initial and final state distribution $\Tilde{\rho}$. The more similar the source and the target domains are, the effectiveness of the SLRL algorithm would be much better. The main aim in the SLRL setting is to maximize $J = \sum_{t=0}^h \gamma^t \mathcal{R}(s_t)$ whilst minimizing the number of steps taken to solve the task.

\begin{figure}[H]
\begin{center}
\includegraphics[width = \textwidth]{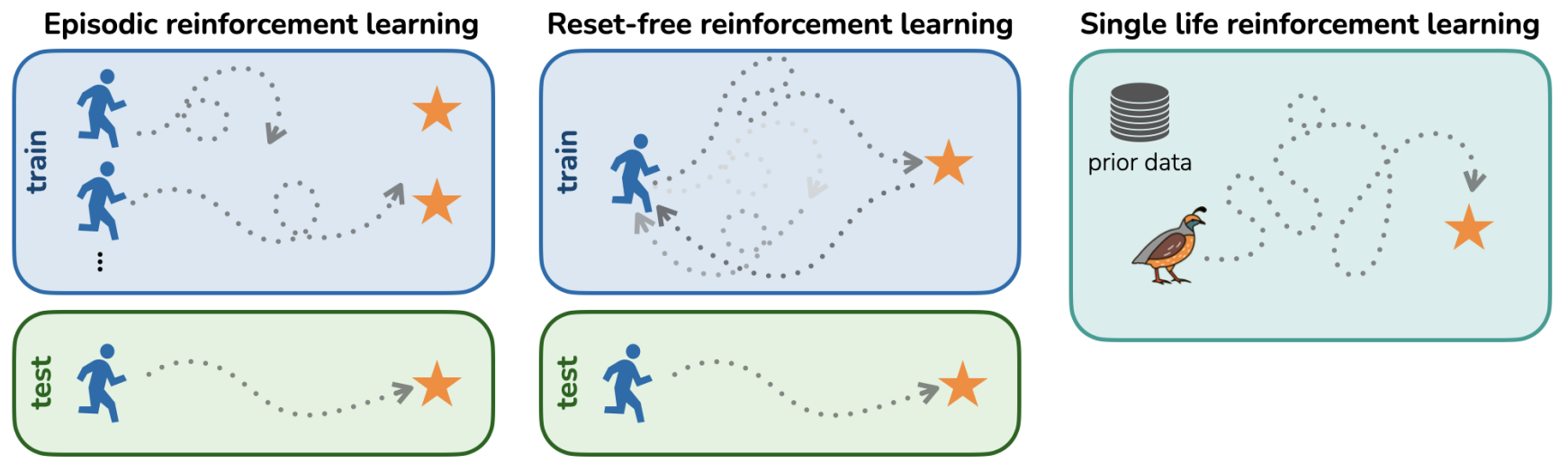}
\end{center}
   \caption{Single Life Reinforcement learning \cite{NEURIPS2022_5ec4e93f}}
\label{fig:1}
\end{figure}
\vspace{-1em}

\subsection{Q-weighted Adversarial Learning (QWALE)}
The desired behavior of the algorithm is to guide the agent toward the distribution of prior data, facilitating recovery from out-of-distribution states and encouraging task completion simultaneously. SLRL assumes access to have sub-optimal offline prior data making it agnostic to the quality of the prior data. The main aim of SLRL is to match the state-action pairs that will lead to task completion rather than matching the entire state-action distribution. By learning a discriminator which can classify whether states belong to prior data or not, the algorithm can down-weigh states which are not in the prior data using the Q values for the state action pairs. Thus, from the paper \cite{NEURIPS2022_5ec4e93f}, Q-weighted adversarial learning (QWALE), minimizes $\mathcal{D}_{\text {JS }}\left(\rho^\pi\| \rho_{\text {target }}^*\right)$ as follows:
$$
\begin{aligned}
& \min_\pi \mathcal{D}_{\text {JS }}\left(\rho^\pi(s, a) \| \rho_{\text {target }}^*(s, a)\right)=\min_\pi \max_D \mathbb{E}_{s, a \sim \rho_{\text {target }}^*}\left[\log D(s, a)\right]+\mathbb{E}_{s, a \sim \rho^\pi}\left[\log (1-D(s, a))\right] \\
& =\min _\pi \max _D \underset{s, a \sim \rho^\beta}{\mathbb{E}}\left[\frac{\exp \left(Q^{\pi_{\text {target }}}(s, a)-V^{\pi_{\text {target }}}(s)\right)}{\mathbb{E}_{\rho^\beta}\left[\exp \left(Q^{\pi_{\text {target }}}(s, a)-V^{\pi_{\text {target }}}(s)\right)\right]} \log D(s, a)\right]+\underset{s, a \sim \rho^\pi}{\mathbb{E}}[\log (1-D(s, a))] \\
& \equiv \min _\pi \max _D \mathbb{E}_{s, a \sim \rho^\beta}\left[\exp \left(Q^{\pi_{\text {target }}}(s, a)-V^{\pi_{\text {target }}}(s)\right) \log D(s, a)\right]+\mathbb{E}_{s, a \sim \rho^\pi}[\log (1-D(s, a))], \\
& \text{where } \mathbb{E}_{\rho^s}\left[\exp \left(Q^{\pi_{\text {target }}}(s, a)-V^{\pi_{\text{target}}}(s)\right)\right] can be ignored.
\end{aligned}
$$

QWALE trains a weighted discriminator (D) function using a fixed Q-function $Q(s,a)$ trained in the source MDP to distinguish between useful transitions and the ones that are less useful. The discriminator outputs essentially act as rewards for the agent (actor) taking actions in the target MDP where +1 represents that the agent's transitions are useful and 0 represents that the transitions were not so useful. The reward for the given state action pair is updated using the negative log likelihood of the probability that the current state does not belong to the prior data. The reward for the given state action pair reduces when the state does not fall closer to the prior data, and using this, the online agent learns the dynamically changing environment to accomplish its newer task.

\section{Methodology}

\subsection{Problem Statement}

\begin{figure}[H]
\begin{center}
\includegraphics[width = 1.1\textwidth]{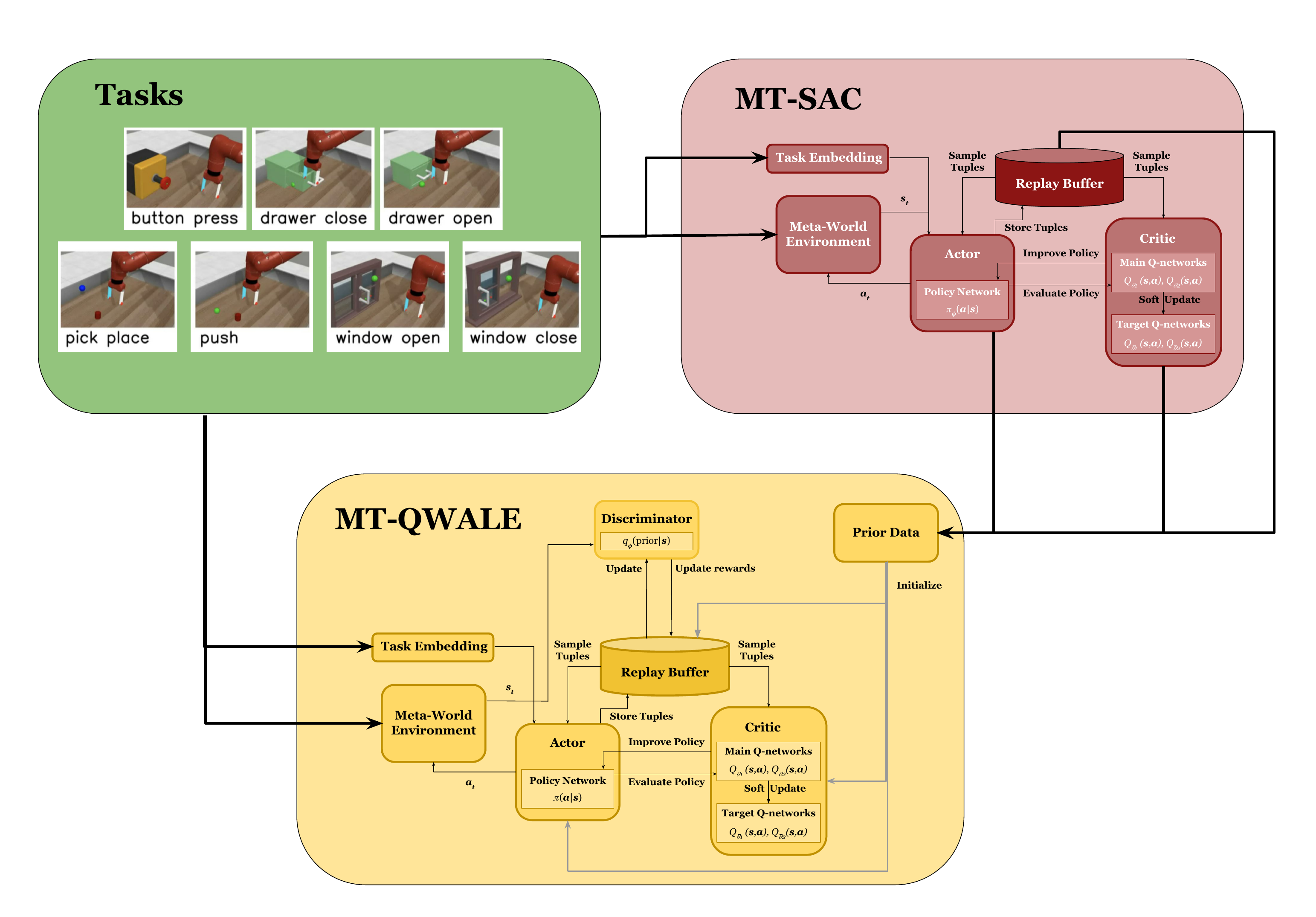}
\end{center}
   \caption{Training Multi-Task QWALE}
\label{fig:4}
\end{figure}

We want to train a robotic arm to be capable of achieving multiple tasks of the MT-10 in a single life. We select seven tasks from the MT-10 (Figure \ref{fig:4}). To do that, we first train an MT-SAC agent on these seven tasks and save the final replay buffer. We will explore different task embedding and select the one that optimizes the performance of the Multi-Task SAC on the seven tasks. Then, we will use this trained policy and replay buffer as the prior data of a Multi-Task QWALE. 

To evaluate the model, we will run the Multi-Task QWALE in the seven tasks environments where the objects are randomly positioned, to introduce novelty in the environments. We will then record the number of successes and the average number of steps to completion and compare it to the performance of MTSAC. Finally, we will do an ablation study by hiding the final goal and observe whether the Multi-Task Agent is still able of achieving the tasks.

\subsection{Multi-Task SAC}

\subsubsection{Model Architecture}
\begin{figure}[H]
\begin{center}
\includegraphics[width = \textwidth]{ 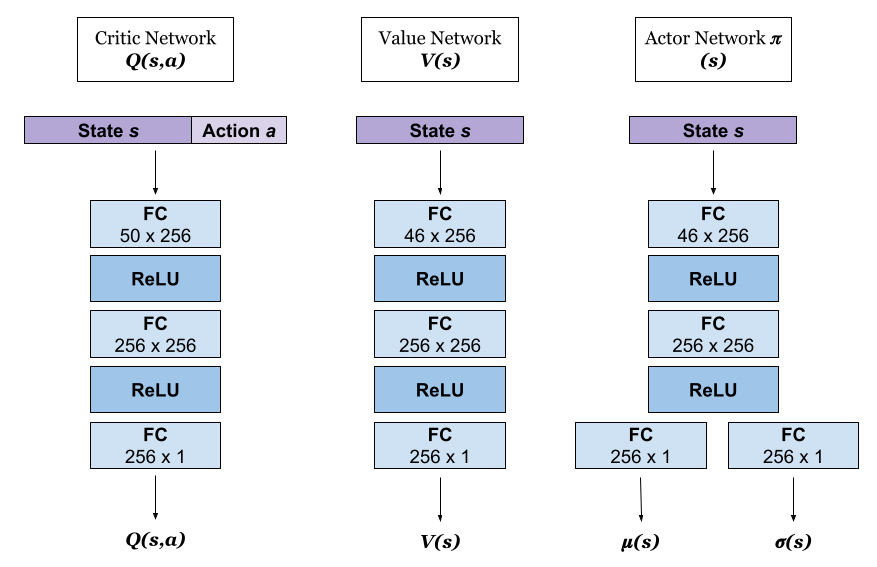}
\end{center}
   \caption{Model Architectures of the Critic, Value, and Actor Networks.}
\label{fig:nets}
\end{figure}

We use multi-task SAC as the algorithm to optimize our multi-task agent. For our critic, value and agent networks we use the architectures shown in Figure \ref{fig:nets}.

The $n_\text{actions} = 4$, since the dimension of the action space of Meta-World is 4. Moreover, the dimension of the state space or observation space is 46 since we concatenate the Meta-World observation (39) with the task embedding (7). For training, this model, $\gamma = 0.99$, $\beta = 0.0003$ and $\alpha = 0.0003$ which are the learning rates for the critic and value networks, and the actor-network respectively. We finally use a reward scale $r_\text{scale} = 2$ and $\tau = 0.005$ for the factor by which we are going to modulate the soft update of our target networks.

\subsubsection{Task Embedding}

We concatenate the observation of the Meta-World with three different types of task embedding: 
\begin{itemize}
    \item \textbf{One-Hot Encoding} which associates each task a unit basis vector $e_k$.
    \item \textbf{Sine Encoding} which associates each task $k$ to the corresponding vector: $\left[\sin(k), \sin(2k), \hdots, \sin(mk)\right]$, given that we have $m$ tasks.
    \item \textbf{Learned Encoding} which associates to a each task $k$ the following task embedding: $M e_k$, where $e_k$ is a unit basis vector and $M$ is a learned matrix. 
\end{itemize}

\subsection{Multi-task QWALE}

We introduce Muti-Task QWALE: Leveraging the Multi-Task SAC algorithm which performs poorly on novel tasks. We tweak the QWALE algorithm to allow the agent to handle a variety of tasks. In this approach, we generate the prior data using the trained MT-SAC algorithm on the different tasks in the 7 environments. For a given task, the Multitask QWALE weights the prior data which contains the same task ID more as compared to the data from the other task. The prior data unrelated to the task is weighted 50\% less than the task-specific data in the prior buffer. The discriminator training remains the same and continues to classify the state as useful followed by weighting the data point using the Q values. Using this, the Multi-Task QWALE allows us to recover from bad states and to keep proceeding toward the goal states based on the prior data.

The discriminator output is used to update the reward for the given state-action pair. The Multi-Task SAC keeps on training online on the observed states using this updated reward function. This allows the model to learn the environment continuously while preferring to visit states which are good if it gets stuck in a bad position or keeps on going to the same states.




\section{Experiments, Results \& Discussion}

We are building a model capable of completing \textbf{multiple tasks} in a \textbf{single trial}, in the presence of \textbf{novel distribution shifts}. In our case, the tasks are defined in Fig. \ref{fig:4}. The novelties are to randomly vary the positions of the objects in the environments within a perimeter of radius 0.3 around the position it was trained on.

\subsection{Multi-Task SAC}
\begin{figure}[H]
\begin{center}
\includegraphics[width = 1.09\textwidth]{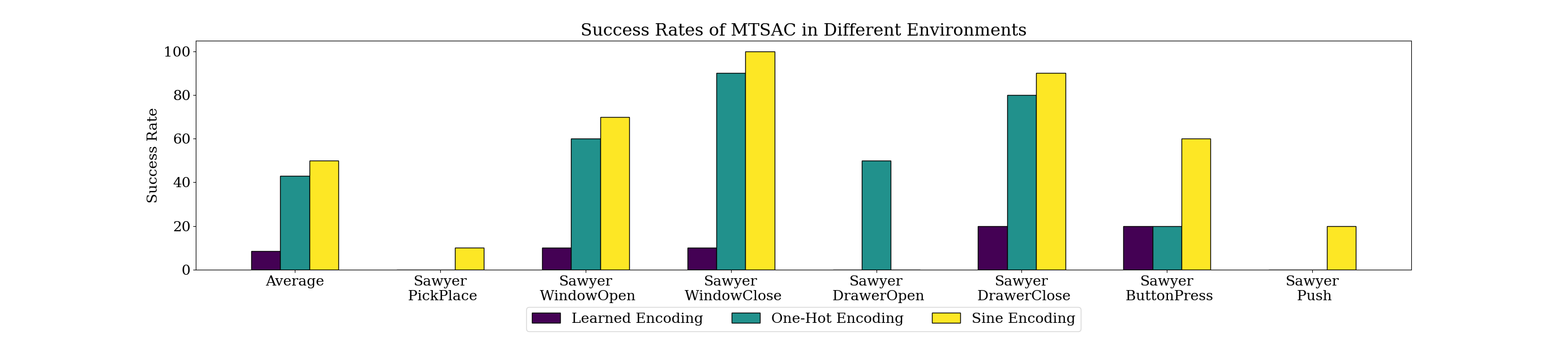}
\end{center}
   \caption{Success Rates of MTSAC in different environments}
\label{fig:5}
\end{figure}

We first train an MT-SAC on the 7 tasks, for 10000 episodes per task to collect the prior data needed for MT-QWALE. We repeat this experiment for each task embedding and select the task embedding that leads to the best performance to use for the Multi-Task QWALE. We observe that using the sine encoding leads to the best performance (Fig. \ref{fig:5}). In fact, using a sine encoding leads to sending more signals to the model, compared to a one-hot encoding, which leads to more shared structure between the tasks while still being linearly independent task embedding. Nonetheless, it is quite surprising to see that the learned embeddings achieved a very low success rate. In fact, we expected the model to learn the best representation of each task that will maximize the performance of the algorithm in each task.

We also observe that the algorithm is less accurate when executing tasks that combine grasping and moving. In fact, the model was able to perform better on simple tasks that require the arm to only move for example Window Close or Drawer Close. On the other side, the model performs poorly on more complex tasks that require the combination of one or more elementary tasks like Pick and Place or Drawer Open which both require some sort of grasping and moving actions.

\subsection{Multi-Task QWALE}

Now that the prior data is collected, we run MT-QWALE in the 7 environments where objects are initialized in random positions. We run each experiment 10 times and collect the success rates of MT-QWALE in each environment as well as the average number of steps to completion. We obtain the following plots:

\vspace{-0.5cm}

\begin{minipage}{0.94\textwidth}
\begin{figure}[H]
\centering
\includegraphics[width = \linewidth]{ 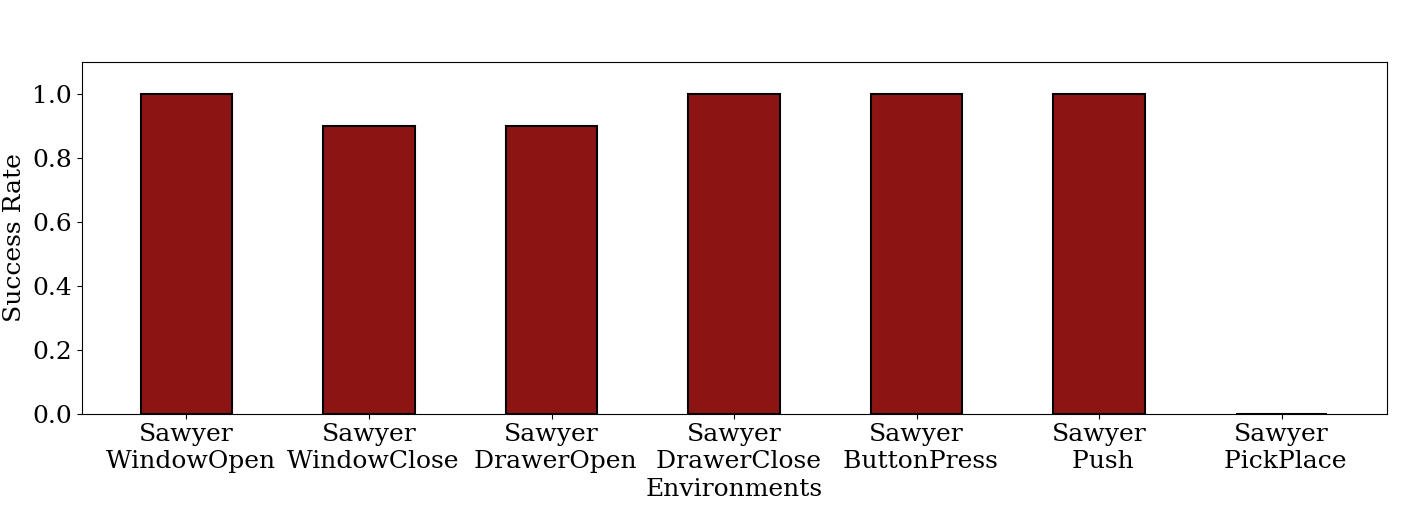}
    \caption{Success Rate of single life RL experiments in different environments}
\vspace{-0.5cm}
\label{fig:srate}
\end{figure}
\end{minipage}
\hfill
\newline
\begin{minipage}{0.94\textwidth}
\begin{figure}[H]
\centering
\includegraphics[width = \linewidth]{ 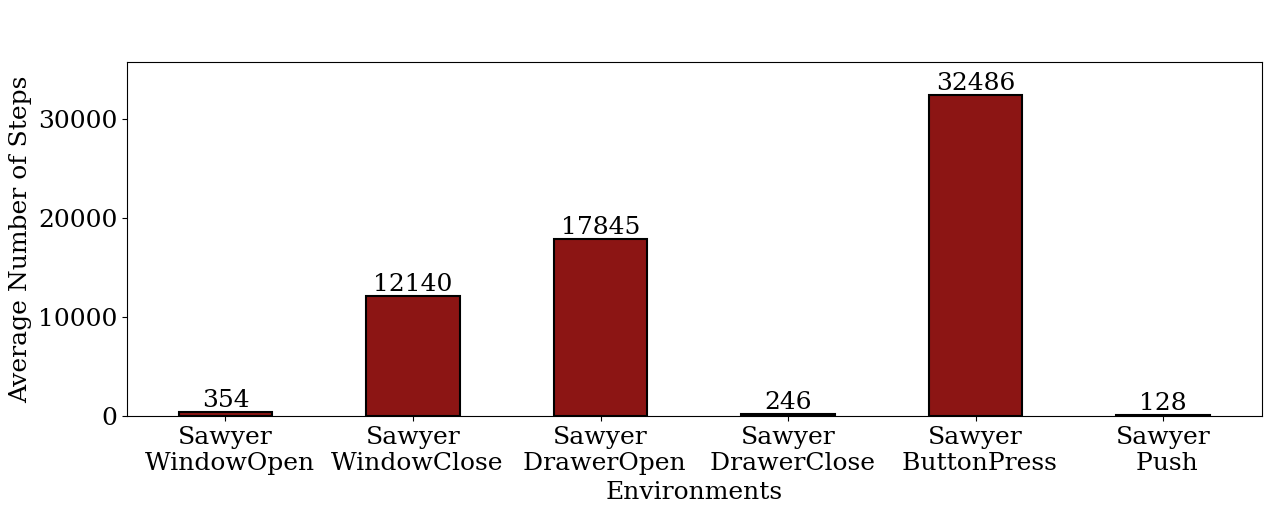}
    \caption{Average number of steps of single life RL experiments in different environments}
\label{fig:steps}
\end{figure}
\end{minipage}

\vspace{0.5cm}

We observe that all the tasks achieve almost 100\% success rate except for the Pick and Place task which failed every single time (Fig. \ref{fig:srate}). Pick and Place is the most complex task in the subset of Meta-World tasks which explains the observed results. It is a combination of moving, grasping, and placing which is much more complex than the other tasks which require just movement and sometimes some grasping.

On the other side, when looking at the average number of steps required to achieve a task there is a close correlation between the number of elementary tasks required to achieve a task and the average number of steps to complete the task. In fact, in Fig. \ref{fig:steps}, simple tasks that require only one to two elementary tasks like Window Open, Drawer Close, or Push require less number of steps to complete the task. We also observe that more complex tasks that require a combination of grasping and moving like Drawer Open need more steps to complete the task. An unexpected result is the Window Close Task, which despite being a simple task requiring only a linear movement of the arm in the opposite direction of Window Open takes much more steps to complete compared to Window Open. The reason behind such behavior might be that the prior data as a whole contains knowledge that is more relevant to solving the task Window Open compared to Window Close. Thus, QWALE would be able to complete the former in fewer steps than the latter.

\subsubsection{Comparing MT-QWALE and MT-SAC in environments with novelty}

\vspace{-0.5cm}

\begin{figure}[H]
    \centering
    \begin{subfigure}[b]{0.30\textwidth}
        \centering
        \includegraphics[width=\textwidth]{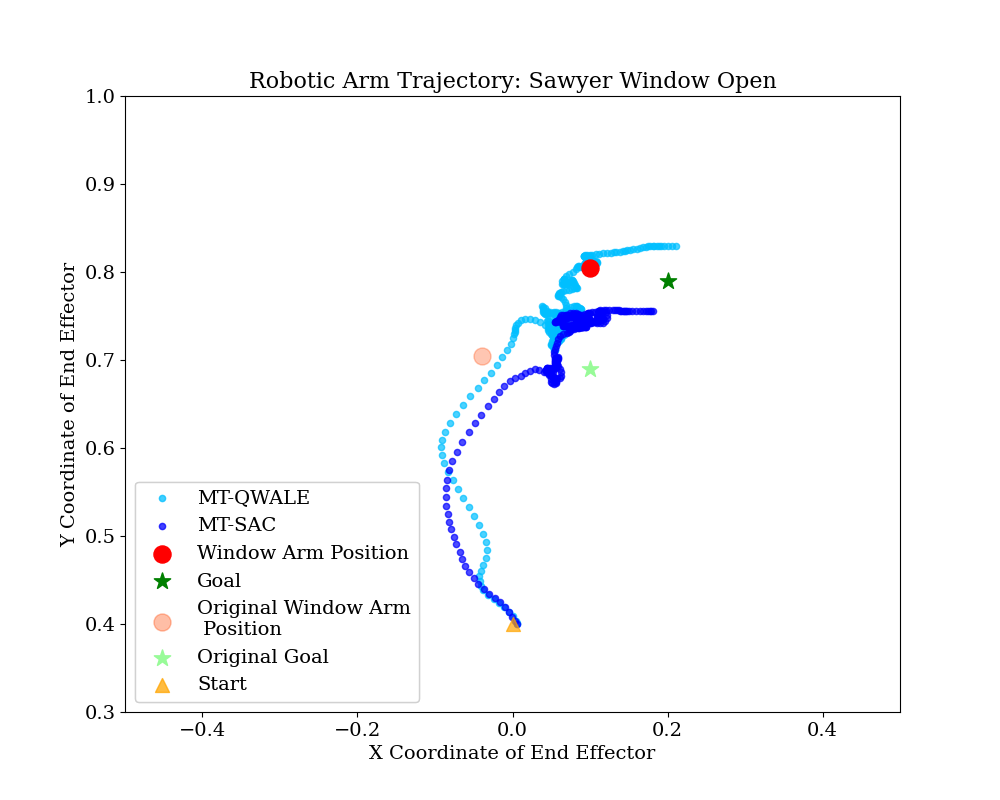}
        \caption{Window Open}
        \label{fig:windowopen}
    \end{subfigure}
    \hfill
    \begin{subfigure}[b]{0.30\textwidth}
        \centering
        \includegraphics[width=\textwidth]{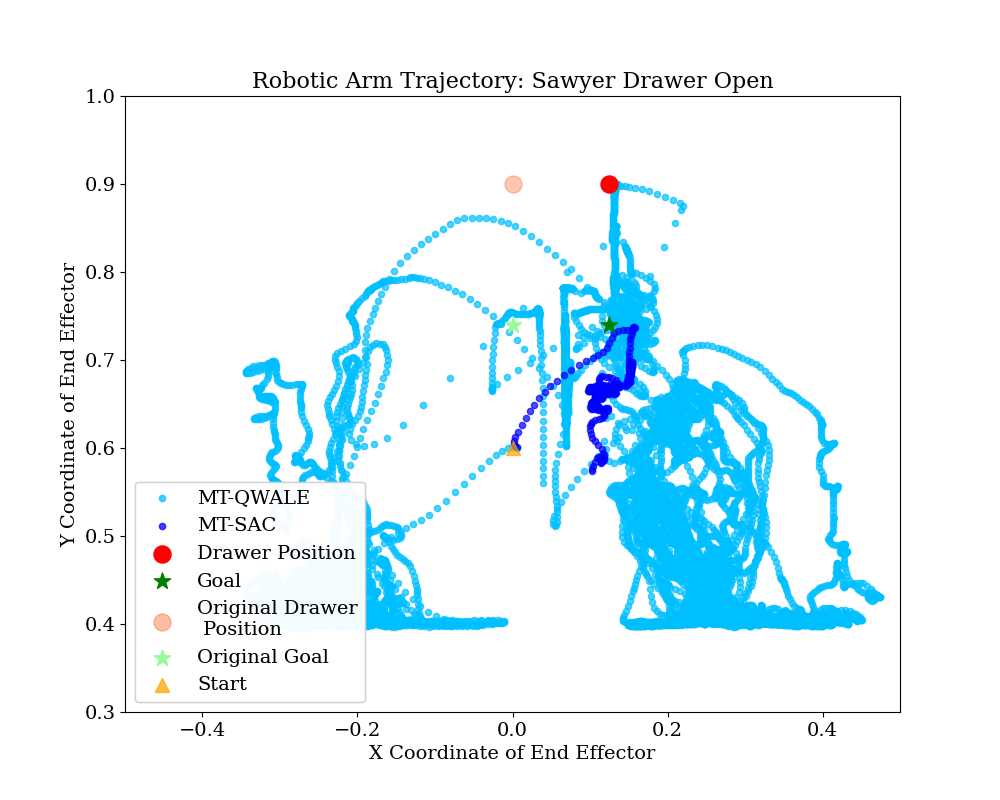}
        \caption{Drawer Open}
        \label{fig:draweropen}
    \end{subfigure}
    \hfill
    \begin{subfigure}[b]{0.30\textwidth}
        \centering
        \includegraphics[width=\textwidth]{ 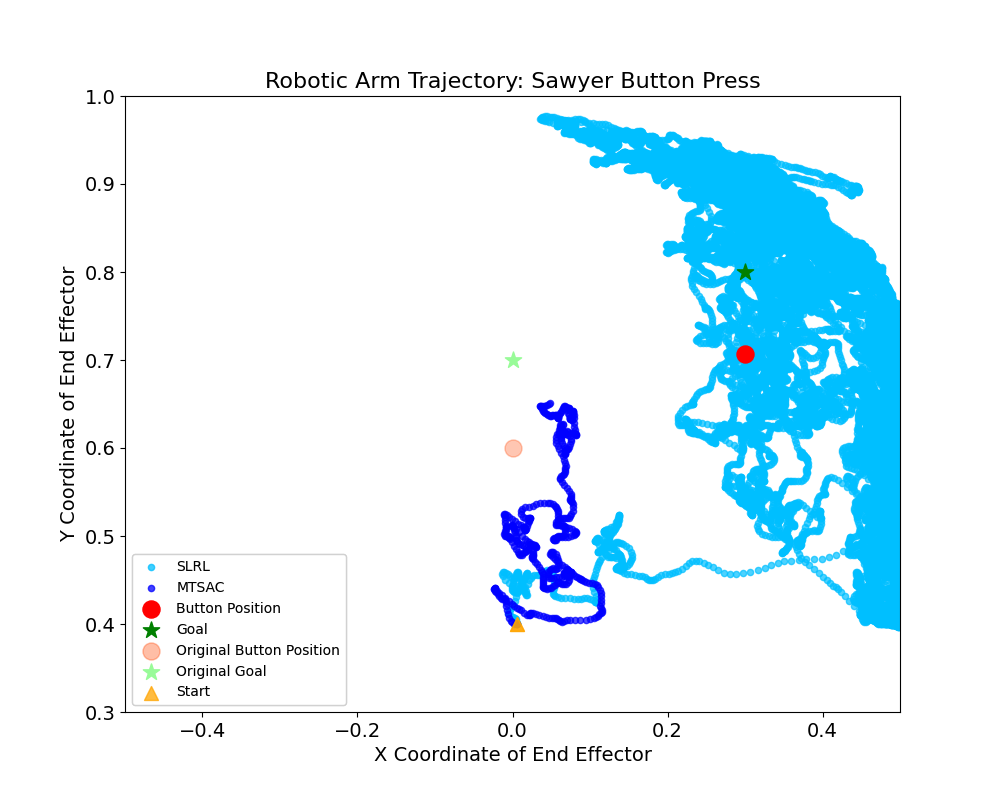}
        \caption{Press Button}
        \label{fig:pressbutton}
    \end{subfigure}
    \hfill
    \begin{subfigure}[b]{0.33\textwidth}
        \centering
        \includegraphics[width=\textwidth]{ 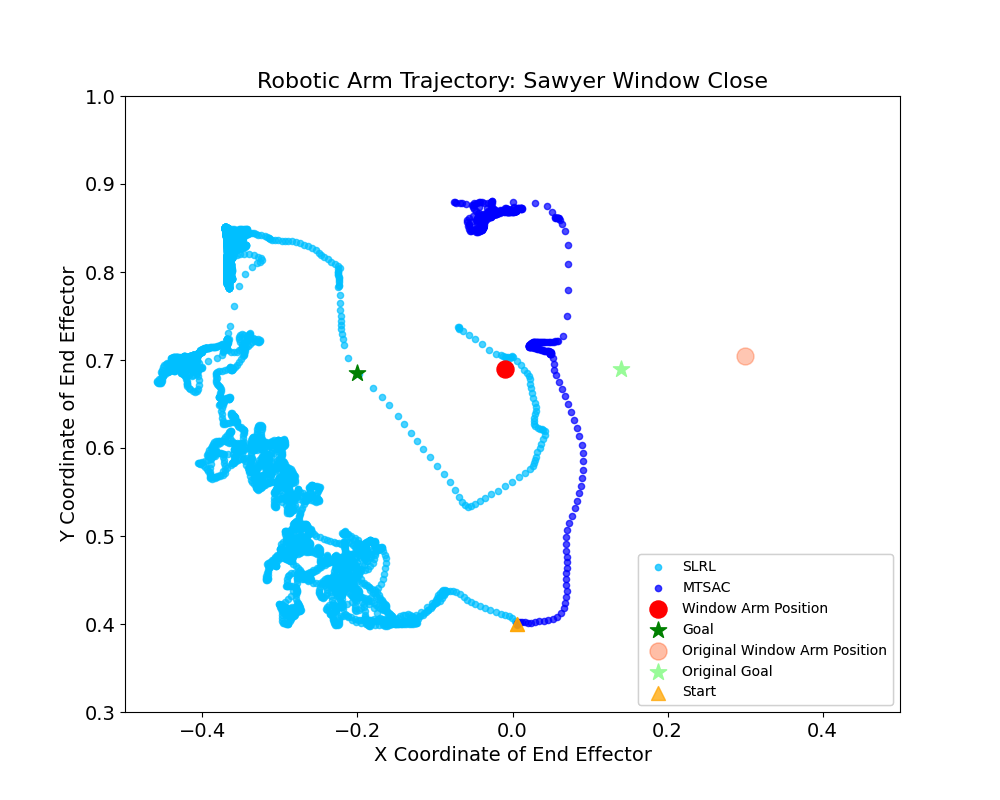}
        \caption{Window Close}
        \label{fig:windowclose}
    \end{subfigure}
    \hfill
    \begin{subfigure}[b]{0.33\textwidth}
        \centering
        \includegraphics[width=\textwidth]{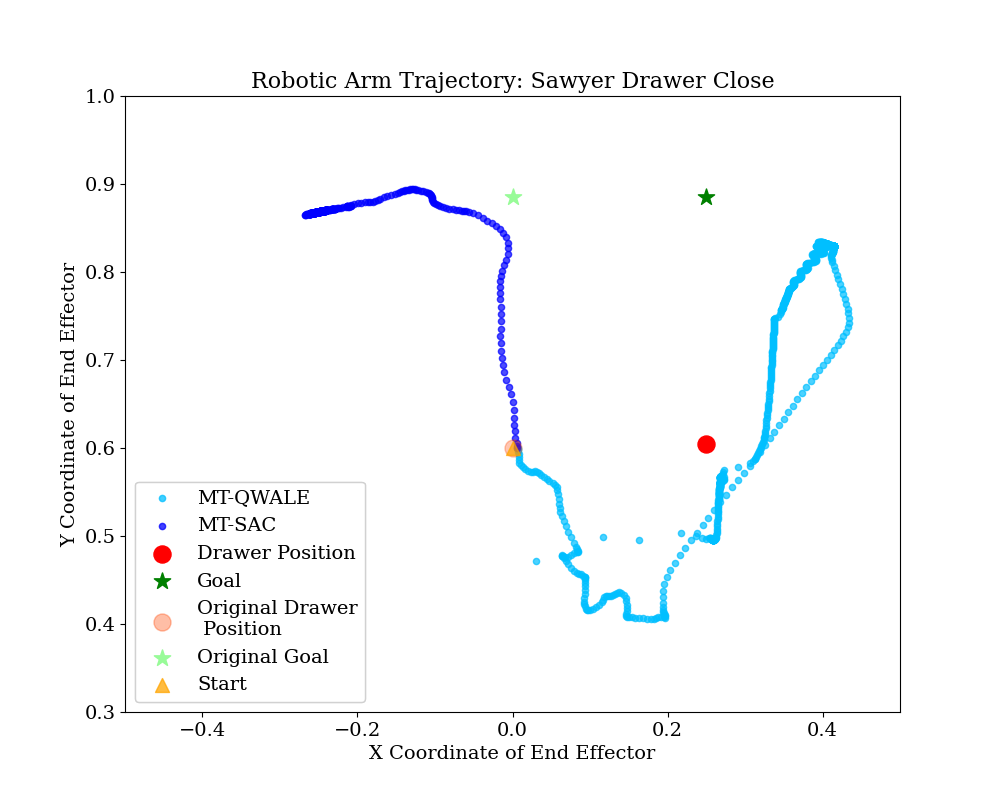}
        \caption{Drawer Close}
        \label{fig:drawerclose}
    \end{subfigure}%
    \hfill
    \begin{subfigure}[b]{0.33\textwidth}
        \centering
        \includegraphics[width=\textwidth]{ 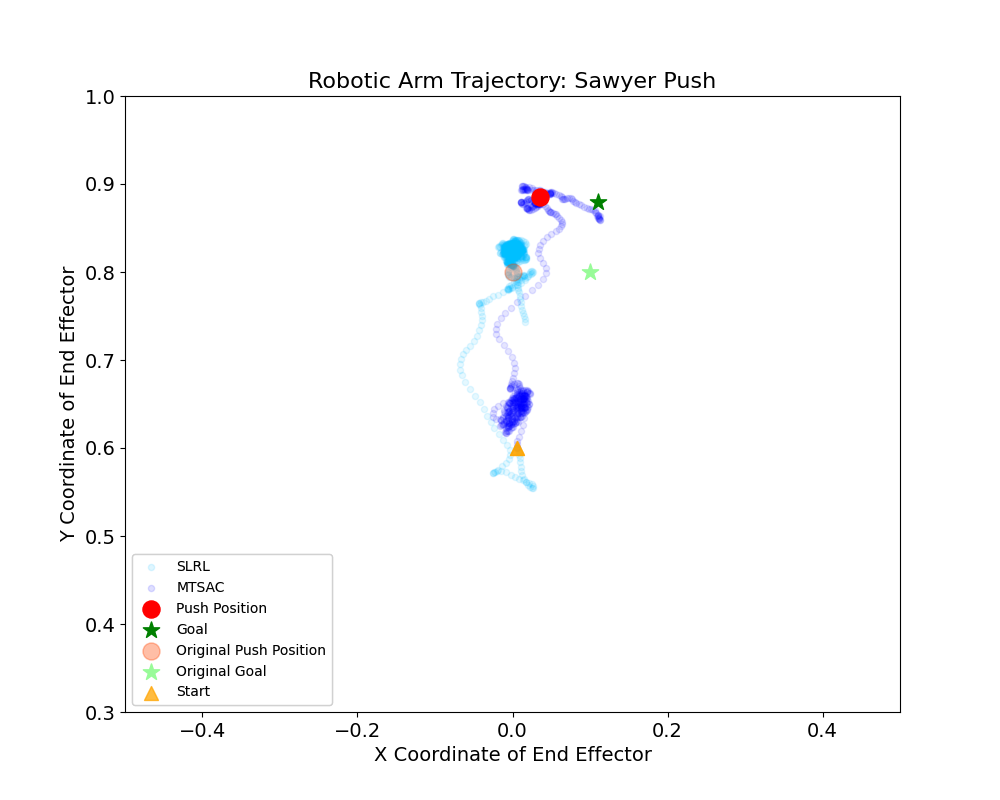}
        \caption{Push}
        \label{fig:push}
    \end{subfigure}
    \caption{Trajectories of end-effector comparing the performance of MT-SAC and MT-QWALE in different environments with novelty}
    \label{fig:grid}
\end{figure}

We observe that MT-QWALE provides a big improvement in the performance of MT-SAC in environments with novelty, except for the Pick and Place task. In fact, in rare cases, we observe that both algorithms achieve the tasks in a similar number of steps as we see in the window open environment (Fig \ref{fig:windowopen}). In the cases where MT-SAC fails in environments with novelty, we observe two different behaviors of MT-QWALE. In some cases, we observe that MT-QWALE achieves the task in a very little number of steps like in the Drawer Close environment (Fig \ref{fig:drawerclose}). This might be the case due to the fact that this is a simple task that requires the arm to just move and thus realize an elementary task. In other cases, we observe that MT-QWALE achieves the task with many steps like in the Drawer Open Environment (Fig (Fig \ref{fig:draweropen})), where the robotic arm had to both grasp the handle and move to achieve this task. This is a more complex task that requires two elementary tasks which may have led to this higher number of steps to achieve the task.

\subsubsection{Ablation Study: MT-QWALE with hidden end goal}

\vspace{-0.7cm}

\begin{minipage}{0.94\textwidth}
\begin{figure}[H]
\centering
\includegraphics[width = \linewidth]{ 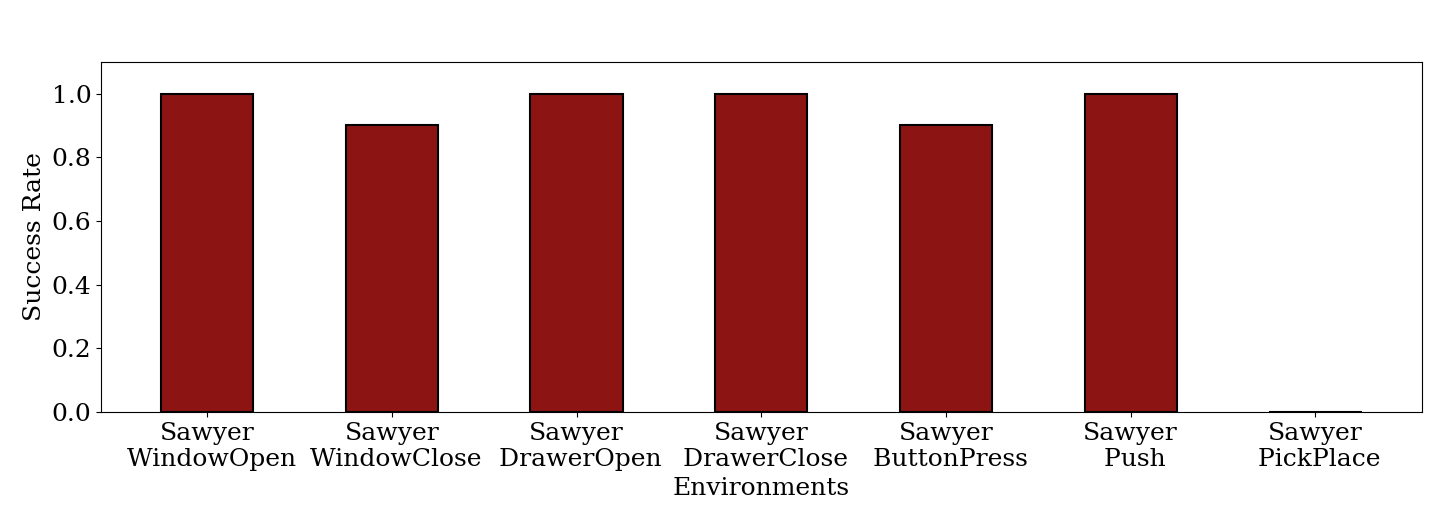}
    \caption{Success Rate of single life RL experiments in different environments with Goal Masking}
\vspace{-0.5cm}
\label{fig:sratemask}
\end{figure}
\end{minipage}
\hfill
\newline
\begin{minipage}{0.94\textwidth}
\begin{figure}[H]
\centering
\includegraphics[width = \linewidth]{ 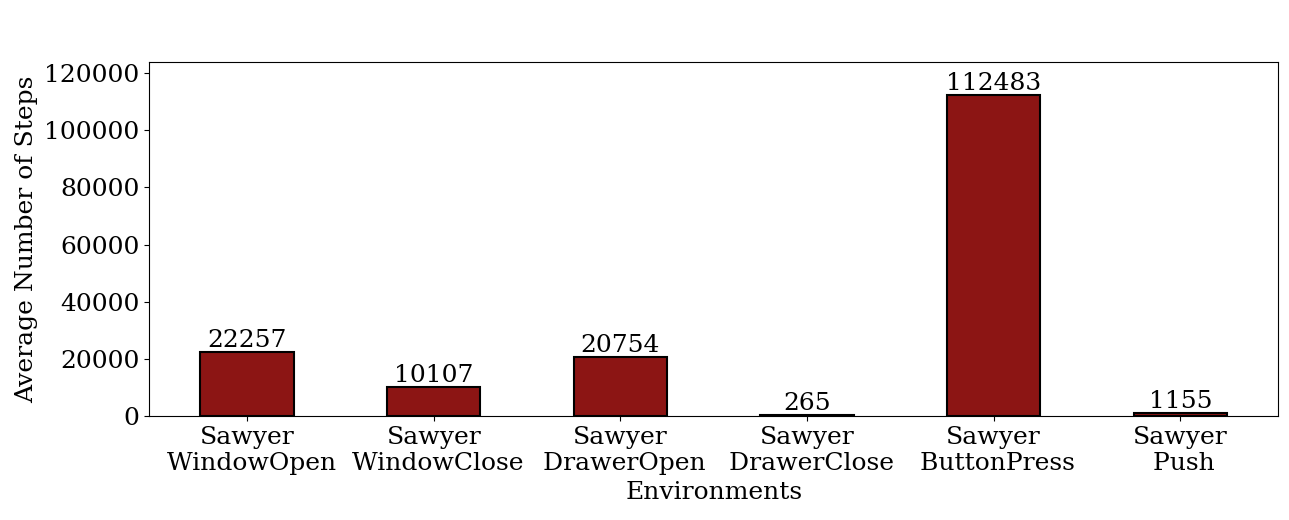}
    \caption{Average number of steps of single life RL experiments in different environments with Goal Masking}
\label{fig:stepsmask}
\end{figure}
\end{minipage}

\vspace{0.5cm}

In this ablation study, we conducted an experiment by concealing the end goal from the observation vector during action selection. Specifically, we assigned zero values to the end goal position. Our aim was to examine how the performance of the MT-QWALE agent is affected by the absence of end goal information.

Interestingly, we found that both masking and not masking the end goal resulted in a comparable success rate. This indicates that the MT-QWALE agent is capable of accomplishing the task without direct knowledge of the end goal, relying solely on reward feedback to adjust its actions. However, we did observe that when the end goal was masked, the average number of steps taken was greater compared to when it was not masked. This outcome aligns with our expectations.

Essentially, by reducing the input signal pertaining to the end goal, we anticipated that the model would make more errors before successfully completing the tasks, thereby necessitating a greater number of steps.

\section{Conclusion}
In conclusion, we present MT-QWALE, a novel approach for multi-task Q-weighted adversarial learning. MT-QWALE builds upon QWALE, adapting it to handle multiple tasks within a single trial. We conduct experiments using seven tasks from Meta-World, namely Window Open and Close, Drawer Open and Close, Pick and Place, Push, and Button Press.

To collect prior data for MT-QWALE, we train MT-SAC on the seven tasks using the best task embedding method we explored, which is sine encoding. We then evaluate the performance of MT-QWALE in these seven environments, incorporating a novelty factor by randomly placing objects within the environment. In comparison to MTSAC, MT-QWALE demonstrates successful completion of all tasks except for the most complex one, Pick and Place.

To further investigate the impact of goal position on MT-QWALE's performance, we conduct an ablation study. We find that hiding the end goal generally leads to an increase in the number of steps required to complete the task, highlighting MT-QWALE's ability to navigate solely based on reward feedback.

For future research, we have several plans in mind. Firstly, we intend to extend this study to include additional tasks and compare the results of MT-QWALE with QWALE using only prior data from a single task. Additionally, we aim to explore different model architectures for the value, actor, and critic networks, as well as investigate alternative methods for task ID embedding. Another avenue of exploration involves conducting ablation studies on other components of the observation vector.

Finally, we are interested in introducing various types of novelties into the environment, such as changing gravity or adding wind, to further enhance the capabilities of our approach. These future endeavors will allow us to deepen our understanding and improve the performance of MT-QWALE in a broader range of scenarios.

\bibliographystyle{unsrt}
\bibliography{references}







\end{document}